\documentclass[runningheads]{llncs}

\usepackage[space,sort,adjust]{cite}
\usepackage{multirow}
\usepackage{graphicx}
\usepackage{color}
\usepackage{subfig}
\usepackage{graphicx}
\usepackage{amsmath,amssymb,amsfonts}

\usepackage{textcomp}
\usepackage{textgreek}
\usepackage{braket,upgreek,bbold}
\usepackage{mathtools}
\usepackage{booktabs}
\graphicspath{{images/}}
\usepackage{mathtools}
\DeclareMathOperator*{\argmax}{argmax} 
\newcommand{\vect}[1]{\boldsymbol{#1}}

\usepackage{array,graphicx}
\usepackage{booktabs}
\usepackage{pifont}

\usepackage{algorithm}
\usepackage{algcompatible}

\begin{document}
\title{MACFE: A Meta-learning and Causality Based Feature Engineering Framework}
\titlerunning{MACFE: Meta-learning and Causality Based Feature Engineering} 
%
\author{Ivan Reyes-Amezcua\inst{1} \and
 Daniel Flores-Araiza\inst{2} \and
 Gilberto Ochoa-Ruiz\inst{2} \and
 Andres Mendez-Vazquez\inst{1} \and
 Eduardo Rodriguez-Tello\inst{3}
 }


%
 \authorrunning{I. Reyes-Amezcua et al.}
%
 \institute{Department of Computer Science, CINVESTAV, México. 
 \and
 Tecnológico de Monterrey, School of Engineering and Sciences, Mexico.
 }

\maketitle         
\begin{abstract}
Feature engineering has become one of the most important steps to improve model prediction performance, and to produce quality datasets. However, this process requires non-trivial domain-knowledge which involves a time-consuming process. Thereby, automating such process has become an active area of research and of interest in industrial applications. In this paper, a novel method, called Meta-learning and Causality Based Feature Engineering (MACFE), is proposed; our method is based on the use of meta-learning, feature distribution encoding, and causality feature selection. In MACFE, meta-learning is used to find the best transformations, then the search is accelerated by pre-selecting ``original" features given their causal relevance. Experimental evaluations on popular classification datasets show that MACFE can improve the prediction performance across eight classifiers, outperforms the current state-of-the-art methods in average by at least 6.54\%, and obtains an improvement of 2.71\% over the best previous works.
\end{abstract}

\section{Introduction}
\label{introduction}
Extracting features from raw data and transforming them into formats that are appropriate for machine learning models is what is known as feature engineering \cite{heaton2016empirical}. This task is usually carried out by a data scientist with good domain knowledge and the data sources of task at hand \cite{nargesian2017learning,zheng2018feature,kuhn2019feature}. Generally, feature engineering entails a daunting manual labor of designing, selecting and evaluating features where even a great intuition is needed \cite{khurana2016cognito,domingos2012few}. This is due to the fact that the performance of most machine learning algorithms heavily relies on the training data quality. This type of datasets usually consists in a large collection of different formats that need to be curated to be exploited by machine learning algorithms \cite{domingos2012few}. Therefore, by using feature engineering, we can select and obtain novel features from the raw data that would better represent the problem.

However, most of the existing automated feature engineering proposals perform this task by applying the \textit{expansion-reduction} method \cite{khurana2018feature}, which is the process of trying a predefined set of transformation functions applied to raw features. Then, those transformed features are selected based on the improvement of model performance or some evaluation metric \cite{nargesian2017learning}. However, \textit{expansion-reduction} leads to an exponential growth in the space of constructed features, which is known as the \textit{feature explosion} problem \cite{chen2019neural}. 
In addition, extracting novel features without a proper and systematic method can lead to an unnecessary increase in the dimensionality of the data, and hence a poor performance on the learning process of the model \cite{blumer1989learnability}. Thus, the \textit{curse of dimensionality} arises \cite{kuo2005lifting}, which is the potential of high-dimensional data to be more complicated to process than low-dimensional data \cite{duda2006pattern}.
\vspace*{-2mm}
\begin{figure*}[!ht]
\begin{center}
\centerline{\includegraphics[width=\textwidth]{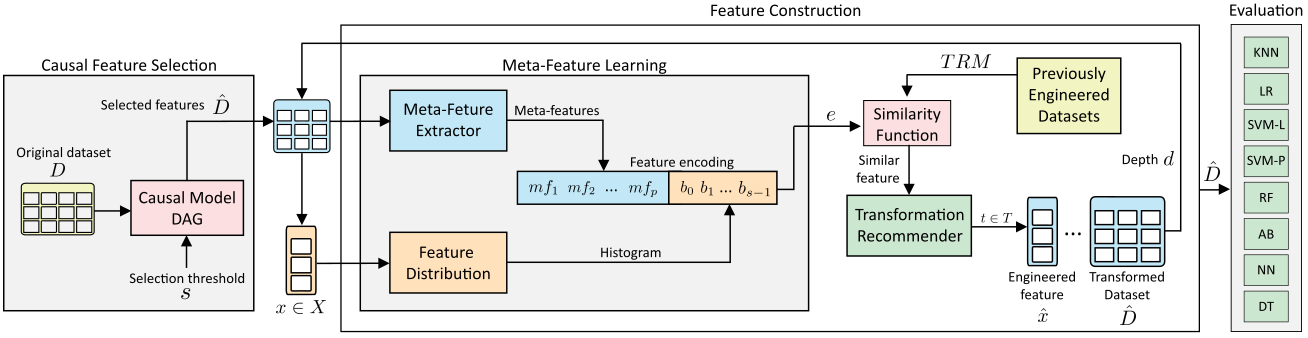}}
\caption{The framework of our method. MACFE extracts meta-features from dataset $\vect{D}$ and a frequency table for each feature $\vect{x} \in \vect{X}$. Then, an encoding $\vect{e}$ is generated by the meta-features and feature distribution. Next, we search for the most similar encoding on the Transformation Recommendation Matrix ($\vect{TRM}$) in order to recommend a useful transformation from it. The transformed dataset $\hat{\vect{D}}$ is built from the constructed novel features and the original ones selected by the Directed Acyclic Graph (DAG) causal model.}
\label{fig:model_overview}
\end{center}
\end{figure*}
\vspace*{-7mm}

It is crucial to realize that there are dozens of types of machine learning models, and each has its own peculiarities and needs \cite{kuhn2019feature}. For instance, some models neither work with highly correlated features nor with highly multi-collinearity. Additionally, other models have troubles dealing with missing, noisy or irrelevant features. Furthermore, since data and models are so diverse, it is difficult to generalize the practice of feature engineering across projects \cite{zheng2018feature}. Thus, finding a proper process to treat data agnostically from a specific learning algorithm can help to choose transformations that better suit the learning process.
To tackle this issue, a possibility is to incorporate only the generated features that have more appropriate knowledge about the data. For this, we present MACFE, a novel meta-learning and causality approach for automated feature engineering for classification problems using tabular data. The main contributions of this paper are briefly described as follows:
\begin{itemize}
\item We present a causality-based method for feature selection on the original dataset.For this, we use the mean magnitude effect of the features on the target to rank and select a subset of them.
\item We propose a novel meta-learning generation for unary, binary and high-order features based on non-linear transformations. This approach addresses the feature explosion problem by only searching on feature transformations that were found useful in past experiences.
\end{itemize}

In order to evaluate the proposed method, we designed a series of experiments on fourteen popular public classification datasets with relatively small dimensions to evaluate the feature generation and selection performance of MACFE. The results are obtained from eight machine learning models: Logistic Regression \textit{(LR)}, K-Nearest Neighbors \textit{(KNN)}, Lineal Support Vector Machine \textit{(SVC-L)}, Polynomial Support Vector Machine \textit{(SVC-P)}, Random Forest \textit{(RF)}, AdaBoost \textit{(AB)}, Multi-layer Perceptron \textit{(MLP)} and Decision Tree \textit{(DT)}. As illustrated in Fig. \ref{fig:model_overview}, our approach is divided into three phases. In the first one, the feature selection is carried out by using a Structural Causal Model (SCM) \cite{reason:Pearl09a} for choosing the most promising features. Then, we move to a meta-learning phase (the second one), where meta-features are extracted from datasets and feature distributions to create encodings for each attribute. Then, we lookup for feature transformation on similar previously engineered datasets. Finally, in the third phase, we evaluate the engineered features among eight machine learning models and obtain the mean accuracy of stratified 5-Fold Cross Validation in order to assess the quality of the feature engineering method. Experimental results show that our proposal is effective on surpassing the scores of state-of-the-art feature engineering methods by achieving a mean accuracy of 81.83\% across the fourteen testing datasets and the eight machine learning models evaluated.

The rest of this paper is organized as follows. In Section 2, we review the state of the art in automated feature engineering. In Section 3, we elaborate on the need of automated methods like ours. In Section 4, we introduce our proposed method MACFE, whereas Section 5 describes the transformation recommender system behind MACFE. In Section 6, we show in detail our evaluation results and finally, in Section 7 the conclusions drawn in this research work are given.

\section{Related Work}

In recent years, many automated feature engineering methods have been proposed using different methodologies. For instance, Data Science Machine (DSM) \cite{kanter2015deep} is an automated feature engineering approach for structured and relational data. DSM proposed a Deep Feature Synthesis (DFS) method, which searches for relations and transformations across features in databases. They include an depth hyper-parameter $d$ for setting the maximum composition, which recursively enumerates  all possible transformations. In addition, DSM generates a large novel feature space, which is reduced by using Singular Value Decomposition (SVD) based feature selection. However, DSM is only suitable for relational data and it could take high computational times due to all the transformation functions used for processing all the original feature sets.

The data-driven approach presented in FCTree \cite{fan2010generalized}, creates novel features from sequential transformations of the original space by employing decision trees and then selecting the best features with the aid of information gain. The method in \cite{piramuthu2009iterative}, known as TFC framework, presents an iterative feature generation algorithm. The method applies  feature transformation across all the features, and then it selects the best features based on information gain. Nevertheless, the generated feature space grows in a combinatorial way, leading to feature explosion. AutoFeat \cite{horn2019autofeat} and AutoLearn \cite{kaul2017autolearn} are also data-driven methods. They can generate large transformations of features, selecting useful features by using regularized regression models for each pair of features. However, these methods require training a regression model, which can be time consuming. Also, they both suffer from the feature explosion problem. Label based Regression (LbR) \cite{wang2020lbr} is another method for generating novel features by using Ridge Regression and Kernel Ridge Regression. This method selects features based on the Distance Correlation Coefficient and the Maximum Information Coefficient (MIC) for each feature pair, which leads to discriminate features that are useful in combination with others.

\subsection{Meta-learning for Feature Engineering}

Recently, meta-learning has been  proposed as a means of improving the quality of the generated features \cite{nargesian2017learning}. Meta-data can be simply defined as data about data \cite{witten2005practical}. For this work, meta-features are used to characterize and identify features with attributes in the context of meta-learning \cite{brazdil2008metalearning,filchenkov2015datasets,rivolli2018towards,alcobacca2020mfe}. Some examples of meta-features are: a) \textit{General}, such as the number of samples, features or classes, etc. b) \textit{Statistical} like standard deviation, correlation coefficients, etc. c) \textit{Information-theoretic} such as entropy, mutual information, noise ratio, etc. d) \textit{Model-based} describes some characteristics of models such as Decision Trees, Bayesian Networks, SVMs, etc.

ExploreKit \cite{katz2016explorekit} is an example of a method that uses meta-learning for ranking and selecting the most promising generated features. ExploreKit does this by applying all possible transformations on features, suffering from the feature explosion problem. Furthermore, Learning Feature Engineering (LFE) is an approach that also uses meta-learning for recommending useful features for classification problems. The transformation recommender in LFE is based on the construction of a meta-feature vector based on the feature values associated with a class label. However, LFE can recommend only unary and binary transformations, lacking high-order transformations.

\subsection{Causality Feature Selection}
Classical feature selection approaches consist in capturing the correlations between features and class variables and lacks from capturing causal relationships among them. In contrast, knowing the causal relationship implies the underlying mechanism of a dataset \cite{yu2020causality}. Hence, basing the feature engineering on relevant causally related features to the class of interest, could provide a richer and more robust output of engineered features.



\section{Problem Definition}
\label{problem}

Let $D = \{\vect{X}, \vect{Y}\}$ be a dataset of input-output pairs, $\vect{X}$ a collection of $n$ features $\{\vect{x_1}, \vect{x_2}, ..., \vect{x_n} \}$, and $m$ labels $\vect{Y} = \{y_1, ... , y_m \}$. A machine learning algorithm $L$ (e.g. SVM, Logistic Regression or Random Forest), and an evaluation metric $E$ (e.g. accuracy, F1-score). 

We refer to a transformation $t \in \vect{T}$ as a function $t(\vect{x})$ that takes a feature as an argument, and maps it to a transformed feature output $\hat{\vect{x}}\in \vect{X'}$. Where $\vect{T}$ is our set of transformations $\{t_1, t_2, ..., t_k\}$ that can be unary or binary, depending on the number of given arguments. Here, a high-order transformation is a composition of unary and binary transformations. Over each feature it is possible to define a series of non-linear transformations, $t_i: \vect{x_i} \rightarrow \mathbb{\hat{\vect{x_i}}}$ that allow to extract as much intra and inter information from the ``original'' data. The goal of feature engineering is thus to transform $\vect{X}$ into $\vect{X'}$ by applying  $\vect{T}$ such that $\vect{X'}$ maximizes the evaluation metric $E$ of a machine learning algorithm $L$. The search for new transformed features and their combinations grows exponentially, and the feature explosion problem arises. MACFE, our proposed feature engineering approach, was devised to help mitigate this problem by employing meta-features to guide the search for transformations on features.

\subsection{Meta-learning and Meta-features}
A formal definition of meta-features was proposed in \cite{rivolli2018towards}, in which meta-features are a set of $q$ values extracted from a dataset $D$ by a function $f$.  \begin{equation}\label{eq:meta-feature}
    f(D) = \sigma(\mu(D, h_\mu), h_\sigma),
\end{equation}

Where $f: D \mapsto \mathbb{R}^q$ is the extraction of $q$ values from dataset $D$, $\mu: D \mapsto \mathbb{R}^{q'}$ is a characterization measure, $\sigma: \mathbb{R}^{q'} \mapsto \mathbb{R}^q$ can be a summarization function such as: mean, minimum, maximum, etc. Moreover, $h_\mu$ and $h_\sigma$ are hyperparameters for $\mu$ and $\sigma$, respectively. Thus, the function $f$ is built by measuring some characteristic from $D$ by $\mu$, and a summarizing function defined by $\sigma$.

Here, meta-features describe features using meta-data. An example is the mean or median, as they are features that provide extra information about the underlying data distribution. In particular, the core of this work is meta-learning applied to the identification of data through meta-features.

\section{Proposed Approach}

\subsection{Datasets}
\subsubsection{Preprocessing.} MACFE is guided by a meta-feature learning based on past experience to create novel features. Our method is trained with $M$ random datasets $\vect{D_{train}} = \{\vect{D_1, D_2, ..., D_M}\}$ collected from Open ML \cite{vanschoren2014openml}, which have a structured format and a classification task related to the data. First, the preprocessing and cleaning of data is performed for each dataset by removing non-numerical features and imputing missing values with the feature mean. Next, a meta-feature \textit{extractor} is used to obtain meta-data about the datasets. Let $\vect{mf}$ be a meta-feature vector composed by the main characteristics of a given dataset $\vect{D_i} \in \vect{D_{train}}$. Thus, a meta-feature vector for a dataset $\vect{D_i}$ is defined as: 
\begin{equation}\label{eq:meta_feature_vector}
\vect{mf} = [mf_1, mf_2, ..., mf_p],
\end{equation}
Where each $mf_i$ is a meta-feature value extracted from the data, and $p$ is the size of the extracted meta-features.
 
However, describing datasets by mapping its main characteristics can be a challenging task. A full set of estimators and metrics can be extracted from a dataset, e.g., the number of classes or instances in a dataset can be a meta-feature value from such a dataset. For this, we use the approach of \cite{pinto2016towards} to perform the automatic meta-feature extraction process. The extraction of meta-features is divided into five categories proposed by Rivolli \textit{et al.} \cite{rivolli2018towards}: simple or general, statistical, information-theoretic and model-based and landmarking. In order to automate the process of extracting meta-features from datasets, we use the framework Meta-feature Extractor (MFE) \cite{alcobacca2020mfe} for each training datasets $\vect{D_i} \in \vect{D_{train}}$, which implements the standard meta-feature extraction described above. 

Next, we treat each feature $\vect{x} \in \vect{D_i}$ as follows: 
\begin{enumerate}
    \item We create a frequency table with a fixed number of \textit{buckets} or \textit{bins} $b$, for each feature $\vect{x}$
    \item A range $r$ is calculated on the feature values given by the $upper$ and $lower$ bounds of the feature.
    \item We generate $s$ disjoint sets or bins $b$ with uniform width $w$. Thus, each bin range $b_i$ is a bucket in which values that are in the bin range lie. Each $b_i$ range starts with the lower bound of $\vect{x}\;plus\;i$ times the width $w$, and ends with the lower bound of $\vect{x}\;plus\;i+1$ times the width $w$.
    \item Finally, each frequency table or histogram is normalized in the range [0,1].
\end{enumerate}

Thus, we obtain an encoding $\vect{e} \in \mathbb{R}^{1\times\eta}$ for each feature $\vect{x} \in \vect{D_i}$, composed by the meta-feature vector $\vect{mf}$ of the dataset and the feature distribution as follows:

\begin{equation}\label{eq:encoding_feature}
    \vect{e}=  [mf_1, mf_2, \dots, mf_p, b_0, b_1, ... b_{s-1}]
\end{equation}

\subsection{Model Training}
\subsubsection{Meta-learning Phase.} The meta-learning phase is described as follows. The unary, binary and scaling feature transformations $t \in \vect{T}$ are applied on the original features $\vect{X}$. Then, an evaluation is performed on both original features and the generated features $t(\vect{X})$. For this, we use the Maximal Information Coefficient (MIC) \cite{reshef2011detecting}, which measures the strength of the linear or non-linear relationships between two variables. MIC generates values between 0 and 1, where 0 means statistical independence and 1 stands for a noiseless statistical relationship between variables. Thus, we get the set of selected transformations $\vect{T_{sel}}$ for each original feature in $\vect{x} \in \vect{X}$ with the maximum score as follows:

\begin{equation}\label{eq:mic_score}
    \vect{T_{sel}} =  \argmax_{t \in \vect{T}}\, g_t\bigg(MIC(t(\vect{x})) - MIC(\vect{x})\bigg)\,.
\end{equation}
Finally, the selected transformations $t \in \vect{T_{sel}}$ are stored in the Transformation Recommendation Matrix ($\vect{TRM}$) for each $\vect{x} \in \vect{D_{train}}$ represented by its corresponding encoding $\vect{e}$. Thus, $\vect{TRM}$ is represented as follows (Fig. \ref{fig:TRM_matrix}). 
\vspace*{-7mm}
\begin{figure}[h]
    \centering
        \[\vect{TRM} = \begin{bmatrix}
        e_{1, 1} & e_{1, 2} & \cdots & e_{1, \eta} & t_1 \\
        \vdots & \vdots & \ddots & \vdots & \vdots\\
        e_{N, 1} & e_{N, 2} & \cdots & e_{N, \eta} & t_N
        \end{bmatrix} \]
    \caption{$\vect{TRM}$ Matrix, where the $i^{th}$ row in the matrix is the feature $\vect{x} \in \vect{D_{train}}$, and the $j^{th}$ column is the encoding value of $\vect{e}$ (Eq. \ref{eq:encoding_feature}). $N$ is the size of all the features in $\vect{D_{train}}$, and $\eta$ is the size of encoding $\vect{e}$ composed by the meta-feature vector $\vect{mf}$ (Eq. \ref{eq:meta_feature_vector}) and feature histogram. The last column stands for the transformations $t \in \vect{T}$ with the resulting highest MIC score for the given features (Eq. \ref{eq:mic_score}).}
    \label{fig:TRM_matrix}
\end{figure}

\vspace*{-15mm}
\begin{minipage}[t]{0.46\textwidth}
  \begin{algorithm}[H]
\caption{Training $\vect{TRM}$}\label{alg:train_t}
\begin{algorithmic}
\STATE {\bfseries Input:} Structured Dataset $D$
\STATE {\bfseries Output:} $\vect{TRM}$

\STATE $D$ = preprocess($D$)

\FOR{each $\vect{x_i} \in D$}
    \STATE $\vect{e_i}$ = encode\_feature($\vect{x_i}$)
    \FOR{each $t \in \vect{T_{un}}$}
        \STATE $\vect{\hat{x}_i}$ = $t(\vect{x_i})$
        \STATE s.append($MIC(\vect{\hat{x}_i})- MIC(\vect{x_i})$)
    \ENDFOR
    \STATE $t_{top} = \argmax(s)$
    \STATE $\vect{TRM}$.append($\vect{e_i}$, $t_{top}$)
    
    \FOR{each $\vect{x_j} \in D | j > i$}
        \STATE $\vect{e_j}$ = encode\_feature($\vect{x_j}$)
        \FOR{each $t \in \vect{T_{bin}}$}
            \STATE $\vect{\hat{x}_{i,j}}$ = $t(\vect{x_i}, \vect{x_j})$
            \STATE $s_i = MIC(\vect{\hat{x}_{i,j}}) - MIC(\vect{x_i})$
            \STATE $s_j = MIC(\vect{\hat{x}_{i,j}}) - MIC(\vect{x_j})$
            \IF {$s_i > 0$ and $s_j > 0$}
                \STATE $\vect{TRM}$.append($\vect{e_i}$, $\vect{e_j}$, $t$)
            \ENDIF
        \ENDFOR
    \ENDFOR
\ENDFOR
\end{algorithmic}
\end{algorithm}
\end{minipage}
\hfill
\begin{minipage}[t]{0.46\textwidth}
  \begin{algorithm}[H]
\caption{Data Transformation}\label{alg:data_transformation}
\begin{algorithmic}
\STATE {\bfseries Input:} $D$, $d$, $s$
\STATE {\bfseries Output:} $\hat{D}$

\STATE $\hat{D}$ = preprocess($D$)
\STATE $\hat{D}$ = causal\_selection($\hat{D}$, $s$)

\FOR{1 to $d$}
    
    \FOR{each $\vect{x_i} \in \hat{D}$}
    
        \STATE $\vect{e_i}$ = encode\_feature($\vect{x_i}$)
        \STATE $t_{un}$ = Similarity($\vect{TRM}$, $\vect{e_i}$)
        \STATE $\vect{\hat{x_i}}$ = $t_{un}(\vect{x_i})$
        \STATE $\hat{D}$.append($\vect{\hat{x_i}}$)
        
        \FOR{each $\vect{x_j} \in \hat{D}$, $\vect{x_i} \neq \vect{x_j}$}
        
            \STATE $\vect{e_j}$ = encode\_feature($\vect{x_j}$)
            \STATE $t_{bin}$=Similarity($\vect{TRM}$,$\vect{e_i}$,$\vect{e_j}$)
            \STATE $\vect{\hat{x_{i,j}}}$ = $t_{bin}(\vect{x_i}, \vect{x_j})$
            \STATE $\hat{D}$.append($\vect{\hat{x_{i,j}}}$)
        
        \ENDFOR
        
    \ENDFOR
        
\ENDFOR

\STATE $e_{\hat{D}}$ = encode\_dataset($\hat{D}$)
\STATE $t_{scaler}$ = Similarity($\vect{TRM}$, $e_{\hat{D}}$)
\STATE $\hat{D}$ = $t_{scaler}(\hat{D})$

\end{algorithmic}
\end{algorithm}
\end{minipage}

\vspace*{3mm}

In Alg. \ref{alg:train_t} the training procedure to learn the most appropriated unary  $\vect{T_{un}}$ and binary $\vect{T_{bin}}$ transformations is presented. This process is done for each feature in a given dataset $\vect{D}$. Similarly, high-order transformations are built by combining several unary or binary transformations one after the other (Alg. \ref{alg:data_transformation}). The order value of the transformation function is related to the number of times a feature is processed by a transformation, e.g., an input feature $\vect{x_1}$ is given as an argument of the $log$ function, so $f_1(\vect{x_1}) = log(\vect{x_1})$. Then, the resulting feature is combined with another feature $\vect{x_2}$, lets say a multiplication, thus, $f_2(f_1(\vect{x_1}), \vect{x_2}) = mult(log(\vect{x_1}), \vect{x_2})$. Finally, the output feature is given to the unary function $square$. Thus, the final transformed feature $\vect{\hat{x}}$ has an order of 3, and can be seen as follows:
\begin{equation}\label{eq:high_order_x}
    \vect{\hat{x}} = f_3(f_2(f_1(\vect{x_1}), \vect{x_2})) = square(mult(log(\vect{x_1}), \vect{x_2})))
\end{equation}
Hence, we look for the underlying information about data through the extraction of more complex features. This gives us the capability of creating novel features from raw features that apparently do not have any predictive power, but in combination with high-order functions can have suitable predictive power for some machine learning models.

\subsubsection{Causal Feature Selection Phase.}\label{sec:transformation_recommender} Once the $\vect{TRM}$ is trained, MACFE is ready to recommend useful transformations for new datasets and features. For this, we start selecting the most promising original features, a causality based feature selection is performed on the features. A DAG Classifier is trained to discover a causal graph from data. For this, we use the implementation of CausalNex \cite{Beaumont_CausalNex_2021}. This graph underlies the causal relationship between features and a target variable. The mean identified causal magnitude effect of the features on the target is used to rank the features. Then, a given threshold hyperparameter $s$ determines the top $k$ selected features. The resulting subset of selected features are processed to obtain an encoding $\vect{e}$ (Eq. \ref{eq:encoding_feature}). 

Then, for a given feature encoding $\vect{e}$, we search for a transformation in $\vect{TRM}$ by retrieving the most similar feature encoding using the \textit{cosine distance} as a similarity measure \cite{singhal2001modern}. We benefit from this measure for ranking the most similar feature-vectors in the range 1.0 for identical feature-vectors and 0.0 for orthogonal feature-vectors \cite{qian2004similarity}.
Next, the most similar feature transformation is applied over the feature. The process is followed by the binary transformations and iterating over the features in the dataset (Alg. \ref{alg:data_transformation}). Furthermore, a depth $d$ hyper-parameter is set to look for the maximum transformation order across \textit{unary} and \textit{binary} functions. Lastly, for the \textit{Scaling transformations} we refer to those transformations on features that changes the scale on a standard range. Many machine learning algorithms struggle to find patterns on data when features are not in the same scale. For this, having scaled features can help gradient descent to converge faster towards a minimum. 

We scale features as follows. For a given feature $\vect{x} \in \vect{X}$, the following scaling functions can be applied. \textit{Normalization}, also called \textit{Min-Max Scaler}, is a method that scales each feature value to the range [0,1]. \textit{Standardization}, this method scales each feature value so that the mean is 0 and the standard deviation is 1. \textit{Robust Scaler}, this scaler is useful when the input feature has a lot of \textit{outliers}. The Robust Scaling is done by calculating the median ($50^{th}$ percentile), and also the $25^{th}$ and $75^{th}$ percentiles. Then, each feature value is subtracted from the median, and divided by the Interquartile Range (IQR) \cite{wan2014estimating}. In order to learn and recommend which scaler is appropriate for a given dataset, we follow a series of data testings. First, we test the features to know the proportion of outliers. If this proportion is larger than a certain threshold $\gamma$, then a Robust Scaler is applied on the features. Secondly, if the data follows a normal distribution, then we use a Standard Scaler. In particular, we use a \textit{Shapiro-Wilk} test \cite{hanusz2016shapiro} to evaluate the normality of data. Then, if the test value is greater than 0.05 we consider the data is normally distributed. Finally, if none of the above tests is true about the data, then we use a Min-Max Scaler on the features. The resulting scaling method is saved in TRM according to the dataset encoding.

\section{Experimental Results}
\subsection{Evaluation Details}
\begin{table}[!h]
\centering
\caption{Statistics of 14 Case Study Datasets}
\renewcommand{\arraystretch}{1.2} 
\setlength\tabcolsep{5pt}
\begin{tabular}{clccc} 
\noalign{\smallskip}
\hline
\textbf{ID} & \textbf{Dataset} & \textbf{Labels} & \textbf{Features} & \textbf{Instances} \\ 
\hline
1 & \textit{Pima Diabetes} & 2 & 8 & 768 \\
2 & \textit{Sonar} & 2 & 60 & 208 \\
3 & \textit{Ionosphere} & 2 & 34 & 351 \\
4 & \textit{Haberman} & 2 & 3 & 306 \\
5 & \textit{Fertility} & 2 & 9 & 100 \\
6 & \textit{Wine} & 3 & 13 & 178 \\
7 & \textit{E.coli} & 8 & 7 & 336 \\
8 & \textit{Abalone} & 29 & 7 & 4177 \\
9 & \textit{Dermatology} & 4 & 34 & 366 \\
10 & \textit{Libras} & 15 & 90 & 360 \\
11 & \textit{Optical} & 10 & 64 & 5620 \\
12 & \textit{Waveform} & 3 & 21 & 5000 \\
13 & \textit{Fourier} & 10 & 76 & 2000 \\
14 & \textit{Pixel} & 10 & 240 & 2000 \\
\hline
\end{tabular}
\label{table:datasets}
\end{table}

The evaluation  of MACFE as an automated feature engineering method is performed on a set of fourteen classification datasets and eight machine learning algorithms commonly cited in the literature \cite{kaul2017autolearn,katz2016explorekit,wang2020lbr}. These datasets are from different areas, such as medical, physical, life and computer science. In addition, these datasets are publicly available in the UCI ML Repository \cite{Dua:2019} and OpenML Repository \cite{vanschoren2014openml}. The main statistics of these datasets are shown in Table \ref{table:datasets}.

\subsection{Implementation Details}
For our experiments, we tested the following learning algorithms: Logistic Regression \textit{(LR)}, K-Nearest Neighbors \textit{(KNN)}, Linear Support Vector Machine \textit{(SVC-L)}, Polynomial Support Vector Machine \textit{(SVC-P)} and Random Forest \textit{(RF)}, AdaBoost \textit{(AB)}, Multi-layer Perceptron \textit{(MLP)} and Decision Tree \textit{(DT)}. The scoring method for the evaluations is the mean accuracy of stratified 5-Fold Cross Validation on each dataset. Same as the state-of-the-art methodology for scoring. Each algorithm is used with scikit-learn \cite{scikit-learn} default parameters. This is because our objective is to enhance the accuracy of a model by improving the data through our automated feature engineering process, MACFE.
\vspace*{-5mm}
\begin{table}[!h]
\caption{Mean accuracy results in 5-fold cross validation among  original datasets (ORIG) and consulted state-of-the-art (TFC\cite{piramuthu2009iterative}, FCTree\cite{fan2010generalized}, ExploreKit\cite{katz2016explorekit}, AutoLearn (AL)\cite{kaul2017autolearn}, LbR\cite{wang2020lbr}) and MACFE (ours). The best performing approach is shown in bold, each dataset is shown with its corresponding ID from Table \ref{table:datasets}.}
\label{table:results}
\resizebox{\textwidth}{!}{%
\begin{tabular}{@{}cccccccccccccccccc@{}}
\noalign{\smallskip}
\toprule
\textbf{D. ID} & \textbf{CLF} & \textbf{ORIG} & \textbf{TFC\cite{piramuthu2009iterative}} & \textbf{FCT\cite{fan2010generalized}} & \textbf{EK\cite{katz2016explorekit}} & \textbf{AL\cite{kaul2017autolearn}} & \textbf{LbR\cite{wang2020lbr}} & \textbf{MACFE} & \textbf{D. ID} & \textbf{CLF} & \textbf{ORIG} & \textbf{TFC\cite{piramuthu2009iterative}} & \textbf{FCT  \cite{fan2010generalized}} & \textbf{EK\cite{katz2016explorekit}} & \textbf{AL\cite{kaul2017autolearn}} & \textbf{LbR\cite{wang2020lbr}} & \textbf{MACFE} \\ \midrule
\multirow{8}{*}{\textbf{1}} & \textbf{KNN} & 71.48 & 72.42 & 73.52 & 73.6 & 68.36 & 72.13 & \textbf{75.12} & \multirow{8}{*}{\textbf{2}} & \textbf{KNN} & 78.35 & 81.48 & 82.70 & 82.4 & 83.19 & \textbf{83.33} & 81.27 \\
 & \textbf{LR} & 76.55 & 75.92 & 76.52 & 73.9 & 74.99 & 71.86 & \textbf{77.47} &  & \textbf{LR} & 77.42 & 78.12 & 78.72 & 78.7 & 79.00 & \textbf{90.47} & 86.05 \\
 & \textbf{SVM-L} & 65.23 & 62.71 & 72.52 & 73.7 & 74.85 & 75.22 & \textbf{77.34} &  & \textbf{SVM-L} & 73.54 & 74.54 & 75.75 & 76.1 & 77.30 & \textbf{90.47} & 86.06 \\
 & \textbf{SVM-P} & 64.89 & 65.71 & 70.52 & 72.6 & 76.32 & \textbf{78.32} & 78.12 &  & \textbf{SVM-P} & 53.36 & 58.41 & 66.44 & 33.6 & 81.71 & 80.95 & \textbf{86.57} \\
 & \textbf{RF} & 75.37 & 72.42 & 73.52 & 74.0 & 73.05 & 72.47 & \textbf{78.12} &  & \textbf{RF} & 73.55 & 81.00 & 82.54 & 47.4 & 77.87 & 76.19 & \textbf{85.62} \\
 & \textbf{AB} & 74.34 & 74.08 & 74.08 & 74.3 & 72.52 & 73.01 & \textbf{76.29} &  & \textbf{AB} & 80.74 & 80.00 & 81.04 & 54.0 & 78.83 & \textbf{85.71} & 83.69 \\
 & \textbf{NN} & 64.32 & 64.12 & 64.22 & 67.3 & 72.39 & 72.50 & \textbf{77.86} &  & \textbf{NN} & 80.30 & 81.07 & 82.00 & 82.4 & 84.09 & 85.71 & \textbf{88.03} \\
 & \textbf{DT} & \textbf{72.38} & 70.23 & 70.46 & 70.9 & 71.05 & 71.12 & 71.74 &  & \textbf{DT} & 75.01 & 74.23 & 74.52 & 75.0 & 75.02 & \textbf{83.33} & 75.53 \\ \cmidrule(lr){2-9} \cmidrule(l){11-18} 
\multirow{8}{*}{\textbf{3}} & \textbf{KNN} & 84.31 & 84.66 & 84.87 & 86.0 & 83.46 & \textbf{92.95} & 89.74 & \multirow{8}{*}{\textbf{4}} & \textbf{KNN} & 71.89 & 70.00 & 71.28 & 72.3 & 68.68 & 70.36 & \textbf{76.14} \\
 & \textbf{LR} & 87.44 & 87.26 & 87.39 & 87.7 & 87.95 & \textbf{95.77} & 93.44 &  & \textbf{LR} & 74.19 & 72.07 & 73.96 & 74.5 & 76.16 & \textbf{76.50} & 74.51 \\
 & \textbf{SVM-L} & 87.44 & 86.71 & 87.78 & 88.0 & 84.30 & 90.14 & \textbf{92.58} &  & \textbf{SVM-L} & 74.18 & 73.97 & 74.18 & 75.4 & 75.82 & \textbf{76.01} & 73.53 \\
 & \textbf{SVM-P} & 64.10 & 70.16 & 71.45 & 72.6 & 74.63 & 78.87 & \textbf{93.72} &  & \textbf{SVM-P} & 74.18 & 73.98 & 74.81 & 75.1 & \textbf{75.52} & \textbf{75.52} & 74.51 \\
 & \textbf{RF} & 93.15 & 91.65 & 93.16 & 94.0 & 92.30 & 91.54 & \textbf{95.44} &  & \textbf{RF} & 68.63 & 68.91 & 69.07 & 70.0 & 65.34 & 70.17 & \textbf{72.22} \\
 & \textbf{AB} & 92.02 & 90.94 & 90.12 & 90.3 & 92.43 & 90.14 & \textbf{94.01} &  & \textbf{AB} & 70.25 & 71.19 & 71.57 & 72.2 & 69.93 & \textbf{73.05} & 71.89 \\
 & \textbf{NN} & 93.14 & 92.45 & 92.13 & 93.6 & 92.29 & \textbf{97.18} & 92.58 &  & \textbf{NN} & 73.19 & 69.02 & 71.19 & 72.2 & 70.91 & 75.02 & \textbf{76.13} \\
 & \textbf{DT} & 88.32 & 87.12 & 88.04 & 88.1 & 88.59 & 88.73 & \textbf{94.87} &  & \textbf{DT} & 66.65 & 66.09 & 66.79 & 67.2 & 66.34 & 67.74 & \textbf{73.86} \\ \cmidrule(lr){2-9} \cmidrule(l){11-18} 
\multirow{8}{*}{\textbf{5}} & \textbf{KNN} & 85.00 & 86.00 & 86.00 & 87.0 & 87.00 & 88.00 & \textbf{88.95} & \multirow{8}{*}{\textbf{6}} & \textbf{KNN} & 67.93 & 74.89 & 79.93 & 83.4 & 93.84 & 95.49 & \textbf{97.22} \\
 & \textbf{LR} & 88.00 & 88.00 & 89.00 & 88.0 & 87.00 & 88.00 & \textbf{89.95} &  & \textbf{LR} & 95.52 & 96.89 & 97.24 & 95.1 & 98.30 & \textbf{99.44} & 98.87 \\
 & \textbf{SVM-L} & 85.00 & 87.00 & 88.00 & 87.0 & 87.00 & 87.00 & \textbf{89.95} &  & \textbf{SVM-L} & 83.03 & 88.14 & 89.94 & 90.8 & 98.31 & \textbf{98.87} & 98.32 \\
 & \textbf{SVM-P} & 88.00 & 87.00 & 87.00 & 88.0 & 88.00 & 88.00 & \textbf{88.89} &  & \textbf{SVM-P} & 96.65 & 96.68 & 96.65 & 92.1 & 92.68 & 94.74 & \textbf{99.43} \\
 & \textbf{RF} & 82.00 & 87.00 & 87.00 & \textbf{90.0} & 84.00 & 88.00 & 89.89 &  & \textbf{RF} & 96.07 & 96.68 & 97.12 & 90.0 & 97.20 & \textbf{98.89} & 97.19 \\
 & \textbf{AB} & 79.00 & 83.00 & 84.00 & 83.0 & 79.00 & 85.00 & \textbf{87.84} &  & \textbf{AB} & 85.82 & 88.12 & \textbf{91.23} & 62.8 & 84.71 & 83.03 & 89.27 \\
 & \textbf{NN} & 88.00 & 88.00 & 88.00 & 88.0 & 85.00 & 88.00 & \textbf{90.00} &  & \textbf{NN} & 42.73 & 46.23 & 49.56 & 64.6 & 97.19 & \textbf{98.87} & 98.32 \\
 & \textbf{DT} & 80.00 & 84.00 & 84.00 & 85.0 & 85.00 & \textbf{88.00} & 87.89 &  & \textbf{DT} & 91.57 & 91.79 & 92.01 & 92.5 & 93.22 & 93.37 & \textbf{95.49} \\ \cmidrule(lr){2-9} \cmidrule(l){11-18} 
\multirow{8}{*}{\textbf{7}} & \textbf{KNN} & 86.59 & \textbf{88.42} & 87.56 & 88.4 & 84.82 & 85.39 & 87.81 & \multirow{8}{*}{\textbf{8}} & \textbf{KNN} & \textbf{23.27} & 21.64 & 22.60 & 23.1 & 22.71 & 21.69 & 22.62 \\
 & \textbf{LR} & 75.88 & 78.23 & 79.24 & 82.8 & 87.19 & 87.19 & \textbf{87.73} &  & \textbf{LR} & 24.61 & 23.69 & 23.60 & 24.8 & 26.50 & 25.25 & \textbf{26.84} \\
 & \textbf{SVM-L} & 85.71 & 85.71 & 85.71 & 86.3 & 86.30 & 86.80 & \textbf{88.87} &  & \textbf{SVM-L} & 25.71 & 25.64 & 25.72 & 25.7 & 26.07 & 25.23 & \textbf{26.57} \\
 & \textbf{SVM-P} & 56.54 & 59.32 & 62.14 & 72.3 & 80.33 & 81.59 & \textbf{93.72} &  & \textbf{SVM-P} & 19.46 & 17.64 & 22.12 & 21.4 & 23.77 & 23.98 & \textbf{26.33} \\
 & \textbf{RF} & 82.73 & 83.46 & 83.76 & 85.1 & 86.59 & 84.80 & \textbf{95.44} &  & \textbf{RF} & 22.91 & 18.78 & 23.02 & 23.2 & 22.21 & 24.15 & \textbf{25.52} \\
 & \textbf{AB} & 62.47 & 63.54 & 64.37 & 65.8 & 65.75 & 63.06 & \textbf{93.44} &  & \textbf{AB} & 20.61 & 19.10 & 19.97 & 21.1 & 20.61 & 21.01 & \textbf{21.45} \\
 & \textbf{NN} & 78.28 & 80.37 & 81.97 & 83.7 & 86.90 & 86.89 & \textbf{92.30} &  & \textbf{NN} & 27.53 & 26.32 & 26.41 & 27.1 & 27.81 & 26.40 & \textbf{28.27} \\
 & \textbf{DT} & 79.74 & 76.32 & 77.67 & 80.3 & 76.40 & 82.11 & \textbf{94.87} &  & \textbf{DT} & 19.27 & 19.00 & 19.13 & 19.3 & 19.41 & 19.42 & \textbf{20.13} \\ \cmidrule(lr){2-9} \cmidrule(l){11-18} 
\multirow{8}{*}{\textbf{9}} & \textbf{KNN} & 89.11 & 90.46 & 92.89 & 91.0 & 96.09 & 96.66 & \textbf{97.82} & \multirow{8}{*}{\textbf{10}} & \textbf{KNN} & 70.00 & 71.00 & 71.18 & 73.7 & 69.44 & 70.27 & \textbf{75.28} \\
 & \textbf{LR} & 97.21 & 97.76 & 97.97 & 97.6 & \textbf{98.61} & 97.77 & 97.81 &  & \textbf{LR} & 60.27 & 64.68 & 67.12 & 71.7 & 70.00 & 68.88 & \textbf{79.72} \\
 & \textbf{SVM-L} & 97.21 & 96.02 & 96.27 & 96.3 & 96.92 & \textbf{98.32} & 97.54 &  & \textbf{SVM-L} & 68.61 & 69.88 & 70.83 & 70.4 & 67.22 & 68.61 & \textbf{82.22} \\
 & \textbf{SVM-P} & 94.41 & 94.00 & 94.12 & 92.0 & 93.56 & \textbf{98.04} & 95.90 &  & \textbf{SVM-P} & 2.22 & 36.68 & 47.97 & 47.8 & 49.44 & 50.13 & \textbf{85.83} \\
 & \textbf{RF} & 96.92 & 96.45 & 96.61 & 95.5 & 95.81 & 98.04 & \textbf{98.09} &  & \textbf{RF} & 71.94 & 72.12 & 73.07 & 77.6 & 70.22 & 72.50 & \textbf{86.11} \\
 & \textbf{AB} & 54.13 & 57.12 & 61.00 & 57.3 & 54.96 & 54.13 & \textbf{75.67} &  & \textbf{AB} & 8.05 & 10.12 & 13.11 & 16.9 & \textbf{18.05} & 14.57 & 15.28 \\
 & \textbf{NN} & 98.04 & 97.13 & 97.22 & 97.7 & \textbf{98.22} & 97.77 & 98.09 &  & \textbf{NN} & 71.66 & 72.35 & 74.24 & 75.7 & 78.33 & \textbf{85.56} & 83.06 \\
 & \textbf{DT} & 95.24 & 95.06 & 94.96 & 95.4 & 94.68 & 96.08 & \textbf{96.45} &  & \textbf{DT} & 62.50 & 62.64 & 63.12 & 63.7 & 65.55 & 65.27 & \textbf{73.06} \\ \cmidrule(lr){2-9} \cmidrule(l){11-18} 
\multirow{8}{*}{\textbf{11}} & \textbf{KNN} & 98.77 & 97.20 & 98.02 & 98.0 & 97.03 & \textbf{99.03} & 98.74 & \multirow{8}{*}{\textbf{12}} & \textbf{KNN} & \textbf{82.48} & 81.28 & 82.00 & 82.1 & 81.14 & 81.54 & 81.44 \\
 & \textbf{LR} & 96.49 & 96.40 & 96.40 & 97.0 & 95.83 & 94.82 & \textbf{97.88} &  & \textbf{LR} & 86.58 & 86.72 & \textbf{87.18} & 86.9 & 85.12 & 87.14 & 86.90 \\
 & \textbf{SVM-L} & 94.89 & 94.12 & 95.17 & 95.1 & 94.01 & 94.71 & \textbf{98.49} &  & \textbf{SVM-L} & 86.90 & 84.54 & 86.23 & 86.9 & 84.40 & \textbf{87.18} & 86.66 \\
 & \textbf{SVM-P} & 99.09 & 99.03 & 99.03 & 99.1 & 96.20 & \textbf{99.21} & 99.06 &  & \textbf{SVM-P} & 81.70 & 81.62 & 82.54 & 80.4 & 85.42 & \textbf{86.18} & 83.84 \\
 & \textbf{RF} & 96.38 & 96.36 & 96.91 & 97.3 & 96.57 & 92.68 & \textbf{98.26} &  & \textbf{RF} & 82.10 & 81.45 & 82.04 & 82.1 & 81.12 & 80.90 & \textbf{86.12} \\
 & \textbf{AB} & 68.65 & 67.62 & 68.35 & 69.7 & 73.78 & \textbf{75.46} & 68.17 &  & \textbf{AB} & 83.62 & 82.54 & 82.84 & 83.0 & \textbf{83.78} & 83.04 & 83.34 \\
 & \textbf{NN} & 98.02 & 95.62 & 95.37 & 96.5 & 96.93 & 96.77 & \textbf{98.33} &  & \textbf{NN} & 85.84 & 82.31 & 3.10 & 84.7 & 83.72 & 83.94 & \textbf{86.26} \\
 & \textbf{DT} & 89.90 & 88.00 & 88.46 & 90.4 & 90.41 & 87.42 & \textbf{91.10} &  & \textbf{DT} & 75.04 & 72.46 & 73.00 & 73.2 & 73.06 & 76.60 & \textbf{78.72} \\ \cmidrule(lr){2-9} \cmidrule(l){11-18} 
\multirow{8}{*}{\textbf{13}} & \textbf{KNN} & 83.85 & 82.17 & 83.82 & \textbf{84.0} & 83.55 & 82.17 & 82.85 & \multirow{8}{*}{\textbf{14}} & \textbf{KNN} & 97.75 & \textbf{98.12} & 97.23 & 98.0 & 97.95 & 97.45 & 97.75 \\
 & \textbf{LR} & 79.45 & 79.97 & 80.00 & 82.2 & 83.15 & \textbf{84.03} & 82.20 &  & \textbf{LR} & 94.35 & 94.22 & 94.28 & 95.5 & 95.75 & 94.95 & \textbf{96.75} \\
 & \textbf{SVM-L} & 81.45 & 81.15 & 82.86 & 82.5 & 83.05 & 83.05 & \textbf{84.40} &  & \textbf{SVM-L} & 92.90 & 92.57 & 93.26 & 94.3 & 94.27 & 93.45 & \textbf{97.70} \\
 & \textbf{SVM-P} & 8.70 & 42.25 & 57.97 & 66.7 & 82.30 & 81.10 & \textbf{85.10} &  & \textbf{SVM-P} & 98.35 & 98.22 & \textbf{98.66} & \textbf{98.7} & 97.25 & \textbf{98.66} & 98.10 \\
 & \textbf{RF} & 79.90 & 78.90 & 79.16 & 80.8 & 79.31 & 81.85 & \textbf{84.45} &  & \textbf{RF} & 95.50 & 94.26 & 95.12 & 96.5 & 94.20 & 95.50 & \textbf{97.60} \\
 & \textbf{AB} & 48.65 & 46.66 & 49.29 & 50.0 & \textbf{50.40} & 48.65 & 43.75 &  & \textbf{AB} & 54.05 & 54.00 & 54.86 & 55.3 & 55.60 & 54.05 & \textbf{65.10} \\
 & \textbf{NN} & 81.90 & 82.34 & 83.12 & 83.4 & 85.50 & \textbf{86.90} & 83.40 &  & \textbf{NN} & 97.15 & 97.15 & 97.15 & 97.2 & 97.15 & \textbf{97.90} & 97.20 \\
 & \textbf{DT} & 74.00 & 74.00 & 74.00 & 74.1 & 74.35 & 74.50 & \textbf{75.40} &  & \textbf{DT} & 87.30 & 86.12 & 86.78 & 86.6 & 87.65 & 87.90 & \textbf{88.55} \\ \cmidrule(lr){2-9} \cmidrule(l){11-18} 
\end{tabular}%
}
\end{table}
\subsection{Comparison with previous works}
\begin{figure*}[!h]
\begin{center}
\centerline{\includegraphics[width=\textwidth]{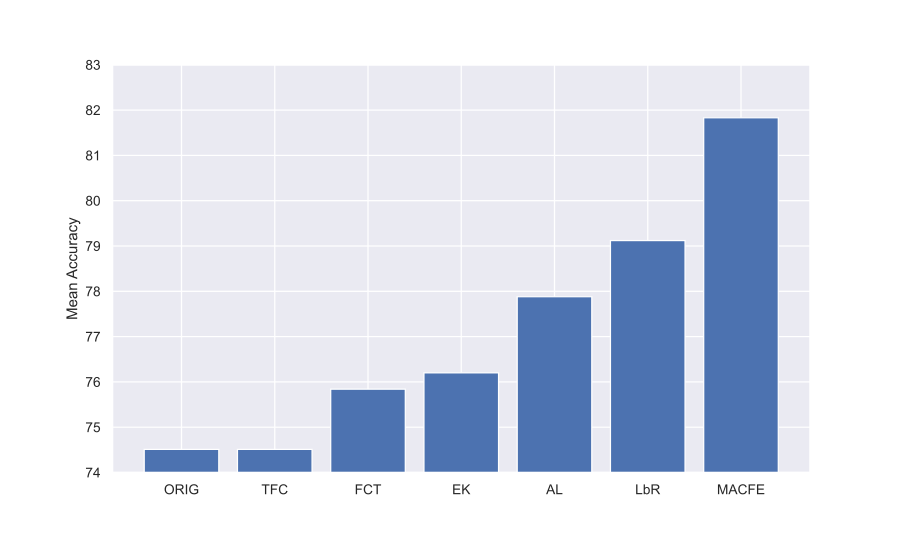}}
\caption{Mean accuracy of state-of-the-art methods and MACFE (ours) across fourteen case study datasets and eight machine learning models.}
\label{fig:macfe_result}
\end{center}
\end{figure*}

The comparison of our proposal takes into account the same scenario conditions of the results presented in recent feature engineering proposals such as TFC\cite{piramuthu2009iterative}, FCTree\cite{fan2010generalized}, ExploreKit\cite{katz2016explorekit}, AutoLearn\cite{kaul2017autolearn} and LbR\cite{wang2020lbr}. In Table \ref{table:results} are shown the scores achieved by our proposal compared against the scores obtained by other approaches in the state-of-the-art. The best scores are shown in bold, each dataset is represented by its ID defined in Table \ref{table:datasets}. 
The improvement among algorithms and datasets is notable: as shown in Fig. \ref{fig:macfe_result} we achieve an average accuracy of 81.83\%  across all tested datasets and classifiers, outperforming TFC, FCT, ExploreKit, AutoLearn (AL), LbR,  by 6.54\%, 5.99\%, 5.63\%, 3.95\%, and 2.71\%, respectively.
\vspace*{-5mm}
\vspace*{-15mm}


\subsection{Discussion}

The transformation recommendation procedure of this method is agnostic from the learning algorithm. But, some transformations can be more appropriate for a certain algorithm. Therefore, MACFE achieves 100\% of efficacy in terms of improving at least one model for each dataset. The depth hyperparameter $d$ of MACFE can generate different orders of complex features to improve the model performance. A high value in $d$ can result in too complex novel features, thus the algorithm cannot learn from the data. In contrast, a small value of the hyperparameter $s$ can lead to a small subset of the original features, thus not finding good relationships between features. Hence, it is recommended a grid search to find the optimal values of hyperparameters.

\section{Conclusions and Future Work}
In this paper, we presented a causality based feature selection to reduce the feature space search for feature transformations. Also, a meta-learning based method for automated feature construction, on which the number of transformations executed on features depends on the number of useful transformations found on historical past similar features. In particular, this method has the capability of constructing novel features from raw data that are informative and useful for a learning algorithm. Hence, MACFE can automatically create features by applying selected transformations to the data, either unary, binary or high-order, instead of applying all possible combinations of those. Hence, the feature explosion problem is minimized. However, MACFE has a fixed set of unary, binary and scaling transformations. In future work, we intend to increase this set by adding more transformation functions, leading to the construction of more informative features from raw features. In addition, the causal selection of features could be improved, since it is applied equally to all datasets but different datasets can be expected to satisfy different causal assumptions, which produces different levels of efficacy when selecting the features to be engineered. To improve this, better methods on general causal discovery are needed.


%
%
%
\bibliographystyle{splncs04}
\bibliography{mybibliography}

\end{document}